\documentclass[lettersize,journal]{IEEEtran}
\usepackage{amsmath,amsfonts}
\usepackage{algorithmic}
\usepackage{algorithm}
\usepackage{array}
\usepackage[caption=false,font=normalsize,labelfont=sf,textfont=sf]{subfig}
\usepackage{textcomp}
\usepackage{stfloats}
\usepackage{url}
\usepackage{verbatim}
\usepackage{graphicx}
\usepackage{cite}

\usepackage{booktabs}
\usepackage{multirow} 
\usepackage{rotating}
\usepackage{bbm}
\usepackage{color}
\usepackage[table]{xcolor}
\usepackage{colortbl}
\usepackage[pagebackref=true,
            breaklinks=true,colorlinks,
            citecolor=cyan,linkcolor=red,
            bookmarks=false]{hyperref}

\begin{document}

\title{Multi-Order Matching Network for Alignment-Free \\Depth Super-Resolution}

% \author{Zhengxue Wang, Zhiqiang Yan, Lina Liu, Xiang Li, Jian Yang}

\author{Zhengxue Wang, Zhiqiang Yan, Yuan Wu, Xiang Li, Guangwei Gao, and Jian Yang
        % <-this % stops a space
\thanks{Zhengxue Wang, Yuan Wu, and Guangwei Gao are with the PCA Lab, School of Computer Science and Engineering, Nanjing University of Science and Technology, Nanjing 210094, China. \protect ~E-mail: \{zxwang, wuyuan, gwgao\}@njust.edu.cn.}
\thanks{Zhiqiang Yan is with the School of Computing, National University of Singapore. \protect ~E-mail: yanzq@nus.edu.sg.}
\thanks{Xiang Li is with the School of Computer Science, Nankai University, Tianjin 300071, China. \protect ~E-mail: xiang.li.implus@nankai.edu.cn.}
\thanks{Jian Yang is with the PCA Lab, School of Computer Science and Engineering, Nanjing University of Science and Technology, Nanjing 210094, China, and also with the PCA Lab, School of Intelligence Science and Technology, Nanjing University, Suzhou 215163, China. \protect ~E-mail: csjyang@njust.edu.cn.}
}


\markboth{Journal of \LaTeX\ Class Files,~Vol.~14, No.~8, August~2021}%
{Shell \MakeLowercase{\textit{et al.}}: A Sample Article Using IEEEtran.cls for IEEE Journals}


\maketitle

\begin{abstract}
Recent guided depth super-resolution methods are premised on the assumption of strict spatial alignment between depth and RGB, achieving high-quality depth reconstruction.
However, in real-world scenarios, the acquisition of strictly aligned RGB-D is hindered by inherent hardware limitations (e.g., physically separate RGB-D sensors) and unavoidable calibration drift induced by mechanical vibrations or temperature variations.
Consequently, existing approaches often suffer inevitable performance degradation when applied to misaligned real-world scenes.
In this paper, we propose the Multi-Order Matching Network (MOMNet), a novel alignment-free framework that adaptively retrieves and selects the most relevant information from misaligned RGB.
Specifically, our method begins with a multi-order matching mechanism, which jointly performs zero-order, first-order, and second-order matching to comprehensively identify RGB information consistent with depth across multi-order feature spaces.
To effectively integrate the retrieved RGB and depth, we further introduce a multi-order aggregation composed of multiple structure detectors. This strategy uses multi-order priors as prompts to facilitate the selective feature transfer from RGB to depth.
%
% Extensive experiments demonstrate that MOMNet achieves state-of-the-art performance and strong robustness. The source codes will be publicly available. 
Extensive experiments demonstrate that MOMNet achieves superior performance and generalization across both unaligned and aligned datasets.
\end{abstract}


\section{Introduction}

\begin{figure}[t]
\centering
\includegraphics[width=0.98\columnwidth]{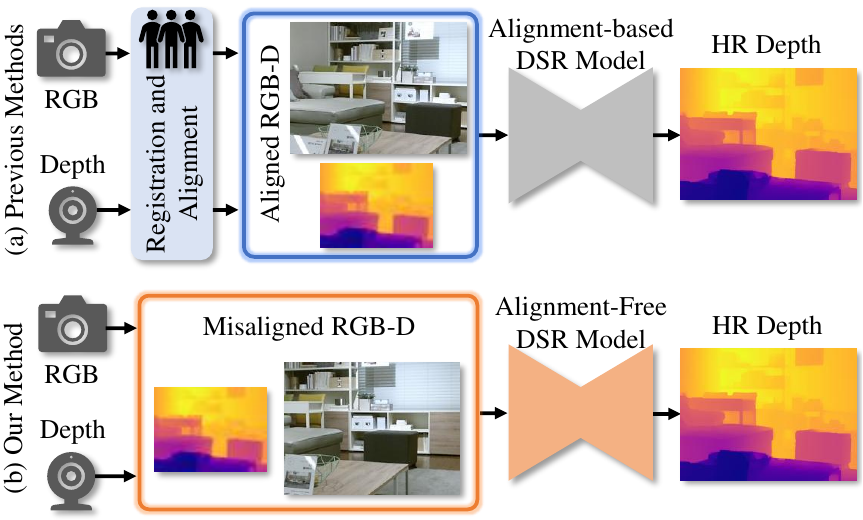}\\
\caption{Previous methods (a) are designed based on the assumption of spatial alignment between RGB and depth data. In contrast, our approach (b) focuses more on addressing the misalignment challenges present in real-world scenarios through multi-order matching, enabling alignment-free DSR.}
\label{fig:fram}
\end{figure}

\IEEEPARstart{D}{epth} super-resolution (DSR) aims to reconstruct high-resolution (HR) depth from degraded low-resolution (LR) counterparts, where RGB is often employed as guidance to enhance depth quality. It is widely applied in various downstream tasks, including 3D reconstruction~\cite{su2019pixel, choe2021volumefusion, wang2023g2, wang2023rgb}, augmented reality~\cite{de2022learning, tang2021joint, liu2018depth}, and virtual reality~\cite{li2020unsupervised, yan2025tri, ye2020depth}.
Recently,  numerous RGB-guided DSR methods~\cite{guo2018hierarchical,ye2025semantics, zhang2025joint,yan2025ducos, zhong2025dual} have been proposed. As illustrated in Fig.~\ref{fig:fram}(a), these approaches are typically premised on the assumption of spatial alignment between RGB and depth, and achieve outstanding performance on aligned RGB-D pairs by developing tailored fusion strategies.

%%%%%%%%%%%%%%%%%%%%%%% Fig. 1 framework --  %%%%%%%%%%%%%%%%%%%%%%%%%%%Previous methods focused on directly integrating RGB and depth features.
%%%%%%%%%%%%%%%%%%%%%%%%%%%%%%%%%%%%%%%%%%%%%%%%%%%%%%%%%%

However, in real-world scenes, obtaining perfectly aligned RGB and depth data is challenging due to hardware limitations and varying imaging conditions. For instance, RGB and depth sensors are typically separate physical modules, making their precise calibration and alignment laborious and costly. Moreover, sensors are often subjected to mechanical vibrations and temperature variations during long-term operation, which can induce calibration drift in the registered camera. This makes it difficult for RGB-D sensors to effectively capture strictly spatially aligned data, particularly in consumer-grade devices with inherently limited registration precision. As a result, previous alignment-based DSR methods are highly susceptible to interference from misaligned RGB structure, which severely degrades the accuracy of the recovered HR depth.
 
\begin{figure}[t]
\centering
\includegraphics[width=0.98\columnwidth]{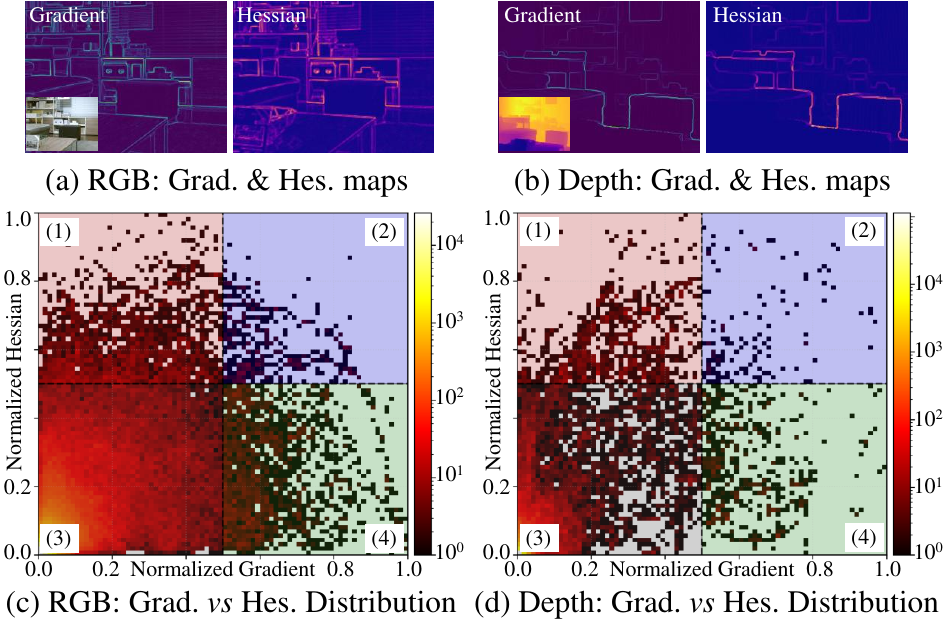}\\
\caption{Visualization of Gradient (Grad.) and Hessian (Hes.) maps for (a) RGB and (b) depth. (c) and (d) are their corresponding distribution comparisons.}\label{fig:Grad_Hessian}
\end{figure}

To address these issues, we propose an alignment-free multi-order matching framework (MOMNet), as shown in Fig.~\ref{fig:fram}(b). This framework adaptively retrieves and aggregates misaligned RGB features within a multi-order feature space, thereby eliminating the dependency on spatially aligned data. MOMNet first introduces a multi-order matching to thoroughly mine the depth-relevant RGB information. This mechanism consists of three key components: zero-order matching for original RGB-D features (without derivative calculation), first-order matching for gradient features (first-order derivative), and second-order matching for Hessian features (second-order derivative). Specifically, due to the spatial inconsistency and inherent modality gap between misaligned RGB and depth, relying solely on original zero-order features is insufficient for accurate matching. To this end, we further incorporate first-order and second-order matching to identify structure-related correspondences in the gradient and Hessian domains, respectively. Figs.~\ref{fig:Grad_Hessian}(a) and (b), the resulting gradient and Hessian maps effectively capture structure details. Furthermore, multi-order distribution comparisons reveal a high similarity between RGB and depth, validating the feasibility of matching retrieval in multi-order space. In particular, Figs.~\ref{fig:Grad_Hessian}(c) and (d) show that gradient and Hessian maps provide complementary information in regions (1) and (4), enabling more comprehensive and robust matching. Subsequently, we design a multi-order aggregation strategy that leverages multiple structure detectors to dynamically integrate the retrieved RGB features. Finally, a multi-order regularization is introduced to facilitate the optimization for alignment-free DSR.

Overall, our contributions are as follows:
\begin{itemize}
    \item We propose a novel alignment-free DSR framework, termed MOMNet, which focuses on addressing the spatial misalignment between RGB and depth in real-world scenes, alleviating the reliance on aligned data. 
    \item We propose multi-order matching, multi-order aggregation, and multi-order regularization strategies. These components operate synergistically within a multi-order space to adaptively retrieve and transfer depth-relevant information from unaligned RGB data into the depth.
    \item Extensive experiments on multiple benchmarks demonstrate the superior robustness and generalization of our method, reaching the state-of-the-art. 
\end{itemize}

\section{Related Work}
\label{sec:rw}

\subsection{Depth Super-Resolution}
Recent advances in DSR have been propelled by numerous methods~\cite{yang2022codon, wang2023learning, chen2024intrinsic, zheng2025decoupling, sun2021learning}, which typically leverage RGB as guidance to reconstruct HR depth. Existing approaches can be broadly categorized into synthetic~\cite{zhong2021high, ye2020pmbanet, wang2026spatiotemporal, zhao2023spherical, tang2021bridgenet} and real-world~\cite{song2020channel, yuan2023structure, he2021towards} DSR. For synthetic DSR, some joint filtering and deep learning methods~\cite{wang2026scene, li2016deep, kim2021deformable} have been developed. For example, DJFR~\cite{li2019joint} employs a Convolutional Neural Network (CNN) to perform joint filtering, which selectively fuses consistent salient structure between the guidance and target images. DKN~\cite{kim2021deformable} integrates a CNN-based architecture to learn sparse and spatially-variant filter kernels, predicting both the relevant neighborhood and corresponding weights for each target pixel. To effectively fuse RGB and depth information, CUNet~\cite{deng2020deep} disentangles cross-modal features into shared and private components via multi-modal convolutional sparse coding. Similarly, DCTNet~\cite{zhao2022discrete} exploits shared and private convolutional kernels to extract common and modality-specific features, respectively. In real-world scenarios, acquired LR depth often suffers from unknown and complex degradation, which poses significant challenges to depth restoration. To address this issue, FDSR~\cite{he2021towards} introduces a real-world RGB-D dataset and establishes a corresponding baseline for real-world DSR. This approach enhances depth by decomposing and fusing high-frequency components of RGB. More recently, DORNet~\cite{wang2025dornet} designs a degradation-oriented DSR framework that learns the implicit degradation representation between LR and HR depth, enabling dynamic depth enhancement. \textit{Unlike previous alignment-based approaches, we propose an alignment-free DSR paradigm that eliminates the dependency on strictly spatially aligned RGB-D data}.

% Additionally, DAGF~\cite{zhong2023deep} captures pixel-level dependencies between RGB and depth by simultaneously predicting dual filtering kernels, thereby effectively transferring information from RGB to depth.

% , which are then fed into a Discrete Cosine Transform to reconstruct HR depth

\subsection{Feature Matching}
Feature matching is a fundamental task in computer vision, aiming to establish point correspondences between images. Recently, a considerable number of feature matching methods~\cite{sarlin2020superglue, jiang2024omniglue} have been proposed. For example, LightGlue~\cite{lindenberger2023lightglue} revisits the sparse matching SuperGlue~\cite{sarlin2020superglue} and proposes a lightweight matching solution. This method predicts a set of correspondences after each computational block, allowing the model to introspect these correspondences and decide whether further computation is necessary. RoMa~\cite{edstedt2024roma} integrates foundational models by designing a Transformer-based matching decoder, significantly enhancing the robustness of feature matching through regression-by-classification. Similarly, OmniGlue~\cite{jiang2024omniglue} leverages vision foundation models as guidance and introduces a keypoint position-guided attention mechanism to disentangle spatial and appearance information. This approach effectively enhances the generalization capability of image matching to unseen real-world scenarios. More recently, MESA~\cite{zhang2024mesa} employs the image understanding capabilities of segmentation foundation models for a region-level matching approach, constructing a multi-relational graph among regions to enhance accuracy. \textit{Different from existing approaches, we propose a novel multi-order matching and aggregation strategy. It synergistically performs zero-order, first-order, and second-order matching within a multi-order space to retrieve RGB information that corresponds to the depth data, thereby effectively enhancing the depth features}.

% we propose a novel multi-order matching and aggregation strategy that simultaneously performs zero-order, first-order, and second-order matching to retrieve RGB information matching the depth within a multi-order space, effectively enhancing depth features.

%%%%%%%%%%%%%%%%%%%%%%% Fig. 2 distribution comparisons  %%%%%%%%%%%%%%%%%%%%%%%%%%%
%%%%%%%%%%%%%%%%%%%%%%%%%%%%%%%%%%%%%%%%%%%%%%%%%%%%%%%%%%

\subsection{Alignment-Free}
 Acquiring strictly spatially aligned guidance and target images is often challenging and time-consuming in real-world applications, prompting the development of alignment-free methods~\cite{han2025hierarchical, li2025progressive}. For instance, DCNet~\cite{tu2022weakly} introduces an alignment-free RGBT salient object detection network, which leverages spatial affine transformation and feature-level affine transformation to model the correlations between RGB and thermal images. Similarly, SACNet~\cite{wang2024alignment} leverages a semantic-guided attention mechanism to learn the relationships between misaligned RGB and thermal images, enabling accurate alignment-free RGBT salient object detection. In addition, IR\&ArF~\cite{qu2025ir} designs a spatial-spectral consistent arbitrary-scale observation network to fuse unregistered hyperspectral and multispectral images. In~\cite{li2024deep}, an alignment-free infrared and visible image fusion network is proposed, which utilizes self-attention to achieve the fusion of unregistered images. \textit{Unlike them, our MOMNet is tailored for RGB-guided DSR, focusing on achieving misaligned and cross-modal fusion between RGB and depth within a multi-order feature space}.

%%%%%%%%%%%%%%%%%%%%%%% Fig. 3 -- Pipeline %%%%%%%%%%%%%%%%%%%%%%%%%%%
\begin{figure*}[t]
\centering
\includegraphics[width=2\columnwidth]{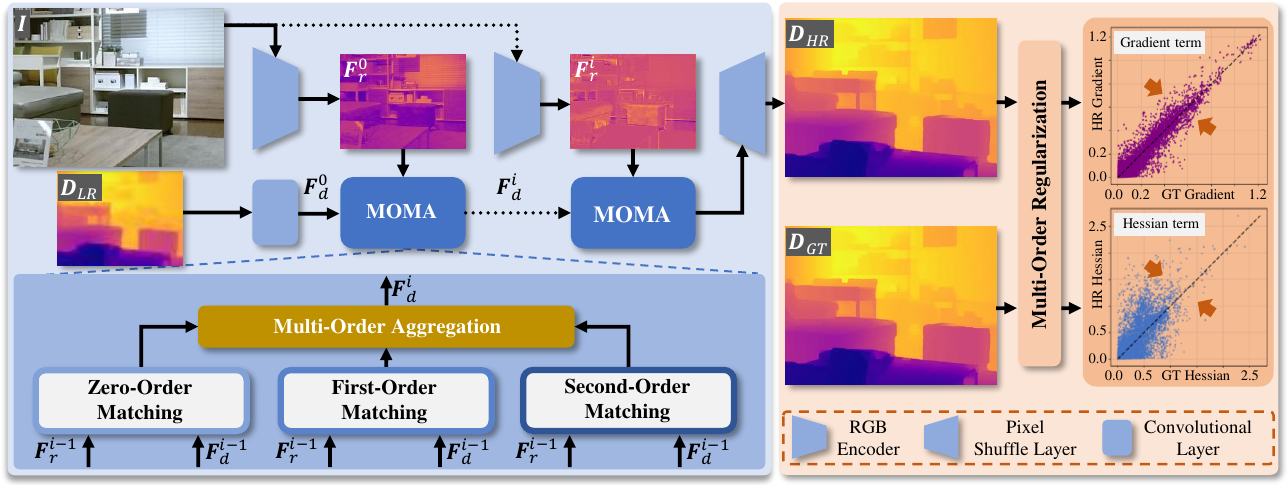}\\
\caption{Overview of MOMNet. Given LR depth $\boldsymbol D_{LR}$ and RGB $\boldsymbol I$ as inputs, we first encode them into features $\boldsymbol F_{d}^{0}$ and $\boldsymbol F_{r}^{0}$. Then, the Multi-Order Matching and Aggregation (MOMA) module is iteratively performed to retrieve and aggregate depth-relevant information from misaligned RGB features, predicting the HR depth $\boldsymbol D_{HR}$. Finally, both $\boldsymbol D_{HR}$ and the ground-truth (GT) depth $\boldsymbol D_{GT}$ are fed into the Multi-Order Regularization to optimize MOMNet.}\label{fig:pipeline}
\end{figure*}
%%%%%%%%%%%%%%%%%%%%%%%%%%%%%%%%%%%%%%%%%%%%%%%%%%%%%%%%%%

\section{Method}
\label{sec:method}

\subsection{Overview}
As shown in Fig.~\ref{fig:pipeline}, given RGB image $\boldsymbol I\in R^{sh\times sw\times 3} $ and LR depth $\boldsymbol D_{LR}\in R^{h\times w\times 1} $ as inputs ($s$ is the upsampling factor. $h$ and $w$ are the height and width of LR depth), our MOMNet first projects them into feature space using RGB encoder (composed of strided convolutions) and $3\times3$ convolutional layers, obtaining initial RGB feature $\boldsymbol F_{r}^{0}$ and depth feature $\boldsymbol F_{d}^{0}$. Then, RGB and depth features are fed into multiple multi-order matching and aggregation (MOMA) modules, which are executed iteratively. These MOMA dynamically retrieve and fuse information corresponding to the depth from the unaligned RGB features, thereby enhancing the depth features. Next, a pixel-shuffle layer is employed to reconstruct the HR depth $\boldsymbol D_{HR}\in R^{sh\times sw\times 1} $ from the enhanced depth features. Given $\boldsymbol D_{HR}$ and GT depth $\boldsymbol D_{GT}\in R^{sh\times sw\times 1} $, we further introduce a multi-order regularization that incorporates both first-order gradient and second-order Hessian terms to constrain the learning of MOMNet in a multi-order space.

To better balance computational cost and performance, we further introduce a lightweight variant, MOMNet-T. By compressing the intermediate feature channels to $0.25\times$ of the original and reducing the MOMA iterations to $2$, this variant retains only $3.35\%$ parameters of the original MOMNet.

\subsection{Multi-Order Matching}
\begin{figure*}[t]
\centering
\includegraphics[width=2\columnwidth]{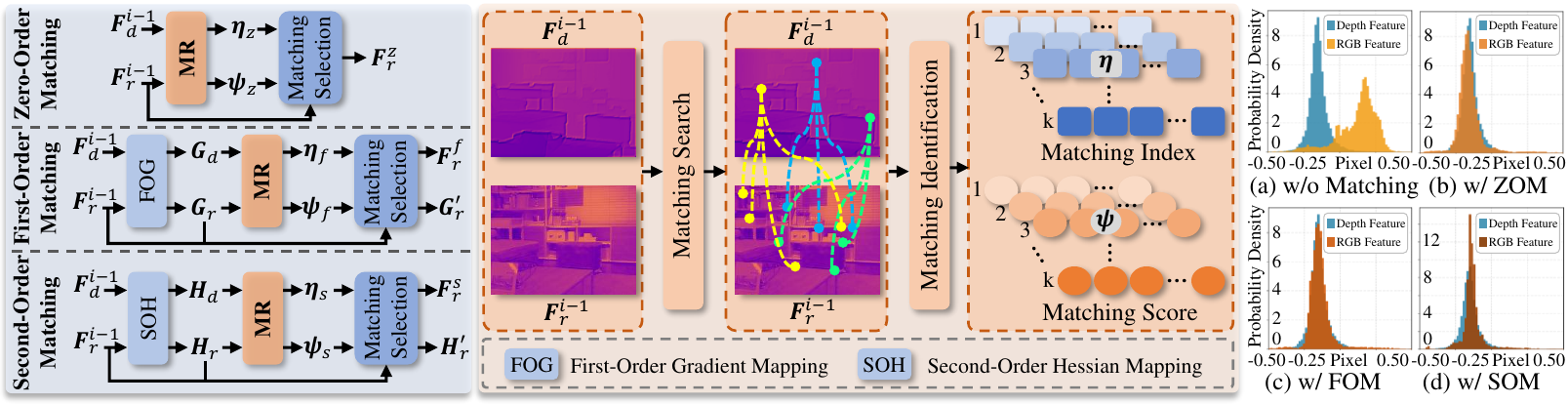}\\
\caption{Details of multi-order matching (left) and matching retrieval (MR, middle). Right: histogram comparison of (a) original RGB, (b) zero-order matched (ZOM) RGB, (c) first-order matched (FOM) RGB, and (d) second-order (SOM) matched RGB.}\label{fig:MOMA_MR_hist}
\end{figure*}

As shown in the blue part of Fig.~\ref{fig:MOMA_MR_hist}, multi-order matching (MOM) is designed to search for RGB information most relevant to depth in multi-order space, consisting of zero-order matching, first-order matching, and second-order matching.

\noindent \textbf{Zero-Order Matching.} As shown in the first row of Fig.~\ref{fig:MOMA_MR_hist} (left), zero-order matching aims to establish the correlation between the original unaligned RGB and depth features. Given the RGB feature $\boldsymbol F_{r}^{i-1}$ and the depth feature $\boldsymbol F_{d}^{i-1}$ as input, we first perform the proposed matching retrieval to compute patch-wise ($3\times3$ patch) correspondence between the RGB and depth features. This process identifies the $k$ most relevant RGB patches for each depth patch, yielding matching index $\boldsymbol \eta _{z}$ and matching score $\boldsymbol \psi_{z}$.

%%%%%%%%%%%%%%%%%%%%%%% Fig. 4 -- distribution comparisons %%%%%%%%%%%%%%%%%%%%%%%%%%%
%%%%%%%%%%%%%%%%%%%%%%%%%%%%%%%%%%%%%%%%%%%%%%%%%%%%%%%%%%

Then, based on the predicted matching indices and scores, we introduce a matching selection strategy that adaptively selects RGB patches matching the depth features from the input RGB features. These patches are aggregated to obtain the zero-order matched RGB features $\boldsymbol F _{r}^{z}$:
\begin{equation}\label{eq:ms}
   \boldsymbol F _{r}^{z}=f_{r} (\sum_{j=1}^{k} f_{p}(\boldsymbol \eta _{z}^{j}, \boldsymbol F _{r}^{i-1})\otimes \tau  (\boldsymbol \psi _{z}^{j})),
\end{equation}
where $f_{p}$ denotes the index-based RGB patch extraction operation. $f_{r}$ refers to a reshape operation that transforms patches into the same dimensions as the RGB feature $\boldsymbol F_{r}^{i-1}$. $\tau (\cdot )$ is softmax function. $\otimes$ is element-wise multiplication.

\noindent \textbf{First-Order Matching.} To mitigate the interference caused by the inherent cross-modal discrepancies in RGB-D data during matching retrieval, we introduce first-order matching for robust retrieval. As depicted in the second row of Fig.~\ref{fig:MOMA_MR_hist} (left), we first calculate the per-channel first-order derivatives of the original RGB feature $\boldsymbol F_{r}^{i-1}$ and depth feature $\boldsymbol F_{d}^{i-1}$, obtaining the RGB gradient representation $\boldsymbol G_{r}$ and depth gradient representation $\boldsymbol G_{d}$. Taking the RGB as an example, the gradient mapping process is formulated as:
\begin{equation}\label{eq:grad}
   \boldsymbol G_{r}=\sqrt{\left (\frac{\partial \boldsymbol F_{r}^{i-1}}{\partial x} \right ) ^{2}+\left (\frac{\partial \boldsymbol F_{r}^{i-1}}{\partial y} \right ) ^{2}} .
\end{equation}

Subsequently, we perform a matching retrieval between RGB gradient $\boldsymbol G_{r}$ and depth gradient $\boldsymbol G_{d}$ in the gradient space, obtaining matching index $\boldsymbol \eta _{f}$ and matching score $\boldsymbol \psi _{f}$. A selection strategy analogous to Eq.\eqref{eq:ms} is applied to extract components from original RGB features that correspond to the depth gradient, yielding first-order matched RGB feature $\boldsymbol F _{r}^{f}$.

Furthermore, we leverage $\boldsymbol \eta _{f}$ and $\boldsymbol \psi _{f}$ to filter the original RGB gradient $\boldsymbol G_{r}$, producing the first-order matched $\boldsymbol G_{r}'$:
\begin{equation}\label{eq:ms_grad}
   \boldsymbol G_{r}'=f_{r} (\sum_{j=1}^{k} f_{p}(\boldsymbol \eta _{f}^{j}, \boldsymbol G _{r})\otimes \tau  (\boldsymbol \psi _{f}^{j})).
\end{equation}
% where $\boldsymbol G_{r}'$ serves as a prompt to guide RGB-D fusion.

%%%%%%%%%%%%%%%%%%%%%%% Fig. 5 -- distribution comparisons %%%%%%%%%%%%%%%%%%%%%%%%%%%
\begin{figure*}[t]
\centering
\includegraphics[width=2\columnwidth]{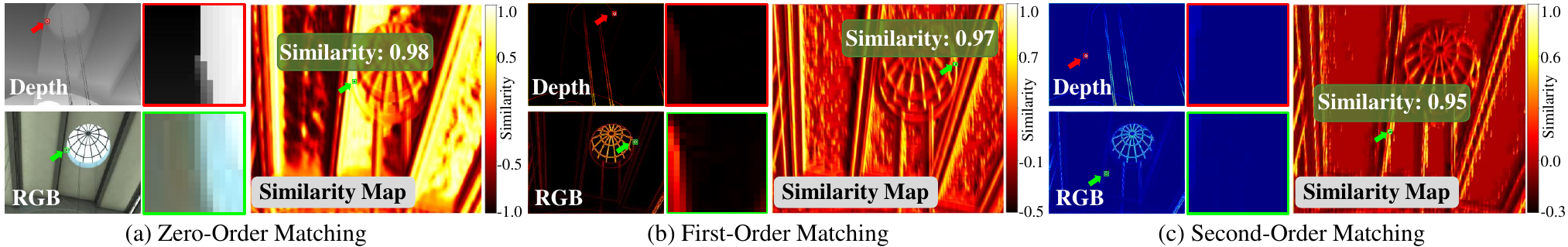}\\
\caption{Visualization of retrieval results. Given a depth patch (red arrow), MOM retrieves all patches in RGB and returns the most relevant one (green arrow).}\label{fig:similarity}
\end{figure*}
%%%%%%%%%%%%%%%%%%%%%%%%%%%%%%%%%%%%%%%%%%%%%%%%%%%%%%%%%%

\noindent \textbf{Second-Order Matching.} Compared to the first-order gradient, the second-order Hessian representation can reveal more intricate local geometric structure. As shown in the third row of Fig.~\ref{fig:MOMA_MR_hist} (left), we further introduce second-order matching for thorough retrieval of high-frequency information. Specifically, given the original RGB feature $\boldsymbol F_{r}^{i-1}$, we first compute the per-channel Hessian matrix and the corresponding Frobenius norm to obtain the RGB Hessian representation $\boldsymbol H_{r}$:
\begin{equation}\label{eq:hessian}
   \boldsymbol H_{r}\!=\!\sqrt{\!\left (\! \frac{\partial^{2} \boldsymbol F_{r}^{i-1}}{\partial x^{2}} \! \right )^{2}\!+\!\left (\! \frac{\partial^{2} \boldsymbol F_{r}^{i-1}}{\partial y^{2}}  \right )^{2}\!+\!2\left ( \!\frac{\partial^{2} \boldsymbol F_{r}^{i-1}}{\partial x \partial y} \! \right )^{2}}.
\end{equation}

Similarly, we apply the same operation as in Eq.~\eqref{eq:hessian} to the original depth feature $\boldsymbol F_{d}^{i-1}$, thereby resulting in depth Hessian representation $\boldsymbol H_{d}$. Then, we conduct matching retrieval to predict the second-order matching index $\boldsymbol \eta _{s}$ and matching score $\boldsymbol \psi _{s}$ between RGB and depth Hessian representations. Finally, $\boldsymbol F_{r}^{i-1}$,  $\boldsymbol H_{d}$, $\boldsymbol \eta _{s}$, and $\boldsymbol \psi _{s}$ are fed together into the matching selection module (analogous to Eqs.~\eqref{eq:ms} and \eqref{eq:ms_grad}) to dynamically generate the second-order matched RGB feature $\boldsymbol F_{r}^{s}$ and RGB Hessian representation $\boldsymbol H_{r}'$.

\noindent \textbf{Matching Retrieval.} As depicted in Fig.~\ref{fig:MOMA_MR_hist} (middle), given RGB and depth features as input, matching retrieval first performs a matching search by computing the correlation between the target depth patch and all RGB patches, resulting in a correlation set $\mathbb{C}\in R^{hw\times hw} $. Taking zero-order matching as an example, the correlation set $\mathbb{C} _{z}$ is defined as:  
\begin{equation}\label{eq:correlation}
   \mathbb{C} _{z}=\phi (\rho (\boldsymbol F_{d}^{i-1}), \rho (\boldsymbol F_{r}^{i-1})),
\end{equation}
where $\rho(\cdot )$ and $\phi(\cdot )$ represent the $3\times3$ patch extraction operation and the cosine similarity function, respectively.

We then retrieve the most relevant RGB information by identifying the top-$k$ patches from the correlation set $\mathbb{C} _{z}$ to enhance the depth representation:
\begin{equation}\label{eq:topk}
    \boldsymbol \eta _{z},\boldsymbol \psi_{z} =topK(\mathbb{C} _{z}),
\end{equation}
where $\boldsymbol \eta _{z} \in R^{hw\times k} $ and $\boldsymbol \psi _{z} \in R^{hw\times k} $ refer to matching indices and matching scores of top-$k$ matched RGB patches.

Fig.~\ref{fig:MOMA_MR_hist} (right) compares the feature distributions of multi-order matched RGB features against depth features. As shown in (a), a significant distribution discrepancy arises from the spatial misalignment between RGB and depth. In contrast, (b)-(d) reveal that zero-order, first-order, and second-order matching strategies effectively retrieve RGB features that are more consistent with the depth representation, thereby substantially reducing the distribution gap. In addition, Fig.~\ref{fig:similarity} presents retrieval examples of our MOM. It is evident that MOM effectively retrieves RGB patches that are highly correlated with the target depth patch from unaligned and cross-modal data. These results further demonstrate the effectiveness and feasibility of our retrieval strategy.

%%%%%%%%%%%%%%%%%%%%%%% Fig. 2 distribution comparisons  %%%%%%%%%%%%%%%%%%%%%%%%%%%
\begin{figure}[t]
\centering
\includegraphics[width=1\columnwidth]{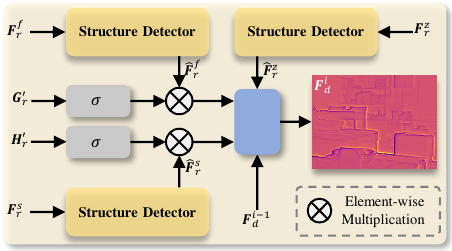}\\
\caption{Details of multi-order aggregation. $\sigma$: Sigmoid Layer.}\label{fig:moa}
\end{figure}
%%%%%%%%%%%%%%%%%%%%%%%%%%%%%%%%%%%%%%%%%%%%%%%%%%%%%%%%%%

Based on the retrieved zero-order, first-order, and second-order RGB features, the multi-order aggregation is designed to selectively transfer RGB information to the depth. As shown in Fig.~\ref{fig:moa}, the multi-order aggregation mainly consists of multiple structure detectors and convolutional layers. Each detector takes $\boldsymbol F_{r}^{z}$, $\boldsymbol F_{r}^{f}$, and $\boldsymbol F_{r}^{s}$ as inputs to extract depth-relevant structure features while suppressing texture noise present in the RGB, generating  $\widehat{\boldsymbol  F} _{r}^{z}$, $\widehat{\boldsymbol  F} _{r}^{f}$, and $\widehat{\boldsymbol  F} _{r}^{s}$.

Next, we utilize the gradient $\boldsymbol G_{r}'$ and Hessian $\boldsymbol H_{r}'$ as prompts to dynamically aggregate the detected RGB structure with the depth feature $\boldsymbol F _{d}^{i-1}$:
\begin{equation}\label{eq:agg}
   \boldsymbol F _{d}^{i}=f_{c}(\cup(\boldsymbol F _{d}^{i-1},\widehat{\boldsymbol  F} _{r}^{z},\sigma (\boldsymbol G_{r}')\otimes \widehat{\boldsymbol  F} _{r}^{f},\sigma (\boldsymbol H_{r}')\otimes \widehat{\boldsymbol  F} _{r}^{s}) ) ,
\end{equation}
where $\boldsymbol F _{d}^{i}$ is the fused depth features, while $f_{c}$, $\cup$, and $\sigma(\cdot)$ denote the convolutional layer, the concatenation operation, and the sigmoid function, respectively.

\noindent \textbf{Structure Detector.} The cross-modal discrepancy between RGB and depth often causes undesirable interference from RGB textures during the fusion process, which hinders depth enhancement. To this end, we introduce a novel learnable structure detector based on the Hessian matrix. 

Drawing on the principles of the Frangi filter~\cite{frangi1998multiscale}, the absolute values and sign of Hessian eigenvalues reveal information about surface curvature and structure type. Specifically, texture-rich corner regions exhibit high curvature in both directions, resulting in two large eigenvalues. For edge or structure, the curvature is high along the normal direction but low along the tangent direction, yielding an eigenvalue pair with a pronounced magnitude disparity. Conversely, flat regions are characterized by two small eigenvalues, where curvature approaches zero in both directions.

\begin{algorithm}
\caption{Structure Detector}
\label{alg:structure_detector}
\begin{algorithmic}[1]
\REQUIRE Input RGB feature $\boldsymbol F_{r} \in \mathbb{R}^{ c \times h \times w}$
\STATE \textbf{Step 1: Initialize Input}
\STATE $\boldsymbol{F}_{norm} \leftarrow \text{Normalize}(\boldsymbol F_{r})$
\STATE $\boldsymbol{F}_{comp} \leftarrow \text{Conv}(\boldsymbol{F}_{norm}, out\_channels=1)$

\STATE \textbf{Step 2: Hessian Computation}
\STATE $\text{Hessian: }\mathcal{H}(\boldsymbol{F}_{comp}(x,y)) =\begin{bmatrix}
\frac{\partial^{2} \boldsymbol F_{comp}}{\partial x^{2}}  &\frac{\partial^{2} \boldsymbol F_{comp}}{\partial x \partial y}  \\
\frac{\partial^{2} \boldsymbol F_{comp}}{\partial y \partial x}  &\frac{\partial^{2} \boldsymbol F_{comp}}{\partial y^{2}}
\end{bmatrix}$

\STATE $\lambda_1, \lambda_2 \leftarrow \text{Eigenvalue Calculation}(\mathcal{H}(\boldsymbol{F}_{comp}(x,y)))$
\STATE Sort eigenvalues: $|\lambda_1| \geq |\lambda_2|$

\STATE \textbf{Step 3: Structure Detection}
\STATE Initialize the learnable parameters $\alpha$ and $\beta$.
\STATE $\text{Structure Awareness: }\boldsymbol{E} \leftarrow \left( 1 - \exp \left( -\frac{|\lambda_1|}{\alpha  + \epsilon }\right)\right)$
\STATE $\text{Texture Suppress: }\boldsymbol{T}_{suppress} \leftarrow \exp\left(-\frac{|\lambda_1 \otimes  \lambda_2|}{\beta + \epsilon}\right)$
\STATE $\text{Geometric Mask: }\boldsymbol{M}_{geom} \leftarrow \mathbb{I}[\lambda_2 < 0]$

\STATE $\text{Structure Descriptor: }\boldsymbol{S} \leftarrow \boldsymbol{E} \otimes \boldsymbol{T}_{suppress} \otimes \boldsymbol{M}_{geom}$

\STATE \textbf{Step 4: Structure Extraction}
\STATE $\widehat{\boldsymbol  F} _{r}=\boldsymbol{S}\otimes \boldsymbol F_{r}$

\RETURN $\text{Structure-enhanced RGB feature } \widehat{\boldsymbol  F} _{r}$
\end{algorithmic}
\end{algorithm}

Based on the above analysis, our structure detector adaptively identifies geometric structure in RGB features by utilizing second-order derivative characteristics, while suppressing texture noise. As outlined in Algorithm~\ref{alg:structure_detector}, the structure detector first performs eigenvalue decomposition to establish a triple-constraint mechanism comprising structure awareness, texture suppression, and geometric mask. Taking RGB features (\textit{e.g.}, $\boldsymbol F _{r}^{z}$, $\boldsymbol F _{r}^{f}$, and $\boldsymbol F _{r}^{s}$) as input, the detector predicts a structure descriptor $\boldsymbol{S}$:
\begin{equation}
\begin{split}
    &\boldsymbol{S}=\underbrace{\left( 1 - \exp \left( -\frac{|\lambda_1|}{\alpha  + \epsilon }\right)\right)}_{\text{structure awareness}} \\
    &\otimes \underbrace{ \exp\left(-\frac{|\lambda_1 \otimes  \lambda_2|}{\beta + \epsilon}\right) }_{\text{texture suppression}} \otimes \underbrace{ \mathbb{I}[\lambda_2 < 0]}_{\text{geometric mask}},
    \label{sd}
\end{split}
\end{equation}
where $\epsilon$ is set to $1\times 10^{-8} $, and $\mathbb{I} [\cdot]$ is the indicator function. $\exp (\cdot)$ denotes the exponential function. Then, $\boldsymbol{S}$ is used to filter input features through a lightweight refinement network (comprising three convolutional layers and a sigmoid layer), yielding the structure-enhanced RGB features.

\subsection{Multi-Order Regularization}
Given GT depth $\boldsymbol D_{GT}$ and predicted HR depth $\boldsymbol D_{HR}$, we introduce a multi-order regularization to optimize MOMNet in a multi-order space. Specifically, we first incorporate a common reconstruction loss $\mathcal{L}_{rec} $ to facilitate depth restoration:
\begin{equation}\label{eq:loss_rec}
   \mathcal{L}_{rec} = {\textstyle \sum_{n\in \mathbb{N} }}  ||\boldsymbol D_{GT}^{n}-\boldsymbol D_{HR}^{n}||_{1},
\end{equation}
where $||\cdot||_{1}$ represents the $L_{1}$ norm, while $\mathbb{N}$ refers to the set of valid pixels in the ground-truth depth.

Then, we further introduce high-order regularization $\mathcal{L}_{hor}$, including first-order gradient term $\mathcal{L}_{grad}$ and second-order Hessian term $\mathcal{L}_{hes}$, to promote the learning of high-frequency components in our MOMNet:
\begin{equation}
\begin{split}
    &\mathcal{L}_{hor}=\underbrace{{\textstyle \sum_{n\in \mathbb{N} }}  ||\Phi (\boldsymbol D_{GT}^{n})\!-\!\Phi (\boldsymbol D_{HR}^{n})||_{1}}_{\text{Gradient term $\mathcal{L}_{grad}$}}  \\
    &+ \alpha \!\cdot\!  \underbrace{{\textstyle \sum_{n\in \mathbb{N} }}  ||\Upsilon  (\boldsymbol D_{GT}^{n})\!-\!\Upsilon  (\boldsymbol D_{HR}^{n})||_{1}}_{\text{Hessian term $\mathcal{L}_{hes}$}},
    \label{loss_grad_hes}
\end{split}
\end{equation}
where $\Phi$ and $\Upsilon$ are gradient mapping and Hessian mapping, as defined in Eqs.~\eqref{eq:grad} and \eqref{eq:hessian}. $\alpha$ is hyper-parameters. The total loss function $\mathcal{L}_{total}$ is defined as:
\begin{equation}\label{eq:loss_grad}
   \mathcal{L}_{total}=\mathcal{L}_{rec}+\mathcal{L}_{hor}.
\end{equation}

%%%%%%%%%%%%%%%%%%%%%%%% Tab. 1 -- Quantitative comparison Bicubic %%%%%%%%%%%%%%%%%%%%%%
\begin{table*}[t]
\caption{Quantitative comparisons between MOMNet and previous methods under misalignment levels of $10\%$, $20\%$, and $30\%$.}\label{tab:qua}
\centering
% \large
% \renewcommand\arraystretch{1.05}
\resizebox{0.94\linewidth}{!}{
\begin{tabular}{l|c|cccccccccccc}
\toprule 
\multirow{2}{*}{Methods} &\multirow{2}{*}{Scale}   &\multicolumn{3}{c}{Hypersim ($\sim30\%$)}   &\multicolumn{3}{c}{DIML ($\sim10\%$)}  &\multicolumn{3}{c}{DyDToF ($\sim20\%$)}   &\multicolumn{3}{c}{Average}\\ %\cline {3-14}
%\cmidrule{2-13}
\cmidrule(lr){3-5}\cmidrule(lr){6-8}\cmidrule(lr){9-11}\cmidrule(lr){12-14} 
 & &RMSE$\downarrow $ &MAE$\downarrow $ &$\delta _{1.05} \uparrow $     &RMSE$\downarrow $ &MAE$\downarrow $ &$\delta _{1.05} \uparrow $  
 &RMSE$\downarrow $ &MAE$\downarrow $ &$\delta _{1.05} \uparrow $     &RMSE$\downarrow $ &MAE$\downarrow $ &$\delta _{1.05} \uparrow $ \\ \midrule

DJF~\cite{li2016deep}                   &\multirow{15}{*}{$\times 4$} &17.08 	&4.07 	&97.67 	        &2.20 	&0.60 	&99.51 	    &5.60 	&1.47 	&98.35 		&8.29 	&2.05 	&98.51       \\
DJFR~\cite{li2019joint}                 &                             &15.68 	&3.33 	&98.15 	        &2.09 	&0.46 	&99.59 	    &4.89 	&1.14 	&98.68 		&7.55 	&1.64 	&98.81       \\
CUNet~\cite{deng2020deep}               &                             &15.36 	&4.29 	&98.07 	        &2.08 	&0.80 	&99.62 	    &4.66 	&1.44 	&98.69 		&7.37 	&2.18 	&98.79       \\
FDKN~\cite{kim2021deformable}           &                             &12.54 	&1.99 	&99.06 	        &2.18 	&0.38 	&99.59 	    &3.44 	&0.62 	&99.33 		&6.05 	&1.00 	&99.33       \\
DKN~\cite{kim2021deformable}            &                             &12.23 	&1.90 	&99.13 	        &2.13 	&0.39 	&99.61 	    &3.43 	&0.63 	&99.33 		&5.93 	&0.97 	&99.36       \\
FDSR~\cite{he2021towards}               &                             &11.46 	&1.69 	&99.22 	        &2.01 	&0.36 	&99.62 	    &3.21 	&0.57 	&99.38 		&5.56 	&0.87 	&99.41       \\
DCTNet~\cite{zhao2022discrete}          &                             &12.50 	&2.91 	&98.71 	        &1.98 	&0.50 	&99.62 	    &3.44 	&0.84 	&99.16 		&5.97 	&1.42 	&99.16       \\
SUFT~\cite{shi2022symmetric}            &                             &9.21 	&0.99 	&99.64 	        &1.85 	&\underline{0.32} 	&99.67 	    &2.75 	&\underline{0.41} 	&\underline{99.53} 		&4.60 	&0.57 	&99.61       \\
DADA~\cite{metzger2023guided}           &                             &15.58 	&3.53 	&97.91 	        &2.24 	&0.55 	&99.54 	    &5.11 	&1.31 	&98.50 		&7.64 	&1.80 	&98.65       \\
SGNet~\cite{wang2024sgnet}              &                             &\underline{8.15} 	&\underline{0.91} 	&\underline{99.66} 	        &1.78 	&0.33 	&\underline{99.68} 	    &2.68 	&0.43 	&99.52 		&\underline{4.20} 	&\underline{0.56} 	&\underline{99.62}       \\
DORNet~\cite{wang2025dornet}            &                             &9.95 	&1.18 	&99.52 	        &1.93 	&0.34 	&99.65 	    &2.80 	&0.43 	&99.51 		&4.89 	&0.65 	&99.56       \\
C2PD~\cite{kang2025c2pd}                &                             &9.60 	&1.12 	&99.55 	        &1.93 	&0.35 	&99.64 	    &2.91 	&0.46 	&99.51 		&4.81 	&0.64 	&99.57       \\
SPFNet~\cite{wang2026scene}             &                             &8.53 	&0.93 	&99.65 	        &1.79 	&\underline{0.32} 	&99.67 	&\underline{2.55} 	&0.47 	&99.50 	&4.29 	&0.57 	&99.61      \\
\rowcolor{gray!30}
\textbf{MOMNet-T}                       &                             &11.14 	&1.67 	&99.15 	        &\textbf{1.66} 	&0.33 	&\textbf{99.70} 	    &3.37 	&0.61 	&99.30 		&5.39 	&0.87 	&99.38       \\
\rowcolor{gray!30}
\textbf{MOMNet}                         &                             &\textbf{7.46} 	&\textbf{0.85} 	&\textbf{99.69} 	        &\underline{1.67} 	&\textbf{0.31} 	&\textbf{99.70} 	    &\textbf{2.41} 	&\textbf{0.38} 	&\textbf{99.56} 		&\textbf{3.85} 	&\textbf{0.51} 	&\textbf{99.65}       \\
\midrule

DJF~\cite{li2016deep}                   &\multirow{15}{*}{$\times 8$} &25.19 	&7.28 	&95.58 	&3.94 	&1.20 	&98.68 	    &8.40 	&2.63 	&96.86 		&12.51 	&3.70 	&97.04      \\
DJFR~\cite{li2019joint}                 &                             &23.06 	&6.17 	&96.36 	&3.43 	&0.95 	&99.00 	    &7.60 	&2.25 	&97.37 		&11.36 	&3.12 	&97.58      \\
CUNet~\cite{deng2020deep}               &                             &21.61 	&5.94 	&96.56 	&3.21 	&0.91 	&99.17 	    &7.05 	&2.09 	&97.58 		&10.62 	&2.98 	&97.77      \\
FDKN~\cite{kim2021deformable}           &                             &19.80 	&4.35 	&97.72 	&3.24 	&0.74 	&99.26 	    &6.29 	&1.52 	&98.24 		&9.78 	&2.20 	&98.41      \\
DKN~\cite{kim2021deformable}            &                             &19.67 	&4.13 	&97.90 	&3.32 	&0.74 	&99.25 	    &6.25 	&1.47 	&98.30 		&9.75 	&2.11 	&98.48      \\
FDSR~\cite{he2021towards}               &                             &18.76 	&3.95 	&97.91 	&2.93 	&0.69 	&99.32 	    &6.04 	&1.45 	&98.32 		&9.24 	&2.03 	&98.52      \\
DCTNet~\cite{zhao2022discrete}          &                             &21.54 	&6.63 	&96.22 	&3.20 	&1.06 	&99.10 	    &6.66 	&2.10 	&97.62 		&10.47 	&3.26 	&97.65      \\
SUFT~\cite{shi2022symmetric}            &                             &15.82 	&\underline{2.40} 	&\underline{98.97} 	&3.03 	&0.65 	&99.32 	    &4.84 	&\underline{0.88} 	&\underline{99.03} 		&7.90 	&\underline{1.31} 	&\underline{99.11}      \\
DADA~\cite{metzger2023guided}           &                             &23.03 	&6.67 	&95.79 	&3.48 	&1.09 	&98.86 	    &7.95 	&2.54 	&96.94 		&11.49 	&3.43 	&97.20      \\
SGNet~\cite{wang2024sgnet}              &                             &15.49 	&2.60 	&98.81 	&2.86 	&\textbf{0.62} 	&\underline{99.37} 	    &4.77 	&0.94 	&98.95 		&7.71 	&1.39 	&99.04      \\
DORNet~\cite{wang2025dornet}            &                             &18.52 	&3.49 	&98.33 	&3.19 	&0.69 	&99.30 	    &5.85 	&1.26 	&98.56 		&9.19 	&1.81 	&98.73      \\
C2PD~\cite{kang2025c2pd}                &                             &\underline{14.83} 	&2.43 	&98.84 	&2.87 	&0.64 	&99.35 	    &\underline{4.74} 	&0.93 	&98.95 		&\underline{7.48} 	&1.33 	&99.05      \\
SPFNet~\cite{wang2026scene}             &                             &16.10 	&2.57 	&98.86 	&2.97 	&\underline{0.63} 	&99.35 	&4.95 	&0.93 	&98.99 	&8.01 	&1.38 	&99.07      \\
\rowcolor{gray!30}
\textbf{MOMNet-T}                       &                             &17.52 	&3.35 	&98.28 	&\textbf{2.76} 	&\underline{0.63} 	&\underline{99.37} 	    &5.47 	&1.21 	&98.55 		&8.58 	&1.73 	&98.73      \\
\rowcolor{gray!30}
\textbf{MOMNet}                         &                             &\textbf{13.91} 	&\textbf{2.09} 	&\textbf{99.06} 	&\underline{2.81} 	&\textbf{0.62} 	&\textbf{99.38} 	    &\textbf{4.41} 	&\textbf{0.83} 	&\textbf{99.09} 		&\textbf{7.04} 	&\textbf{1.18} 	&\textbf{99.18}      \\
\midrule

DJF~\cite{li2016deep}                   &\multirow{15}{*}{$\times 16$}&37.41 	&14.63 	&89.59 	    &7.31 	&2.99 	&95.33 	    &12.78 	&5.43	&92.99		&19.17 	&7.68 	&92.64        \\
DJFR~\cite{li2019joint}                 &                             &36.13 	&13.22 	&91.06 	    &7.01 	&2.60 	&96.29 	    &12.30 	&4.80	&94.10		&18.48 	&6.87 	&93.82        \\
CUNet~\cite{deng2020deep}               &                             &32.19 	&11.76 	&92.17 	    &5.66 	&2.15 	&97.32 	    &11.11 	&4.25	&95.09		&16.32 	&6.05 	&94.86        \\
FDKN~\cite{kim2021deformable}           &                             &30.96 	&9.89 	&93.86 	    &5.61 	&1.88 	&97.65 	    &10.70 	&3.67	&95.77		&15.76 	&5.15 	&95.76        \\
DKN~\cite{kim2021deformable}            &                             &30.70 	&9.57 	&94.16 	    &5.67 	&1.89 	&97.61 	    &10.57 	&3.59	&95.88		&15.65 	&5.02 	&95.88        \\
FDSR~\cite{he2021towards}               &                             &28.36 	&8.63 	&94.72 	    &4.99 	&1.63 	&98.03 	    &9.85 	&3.24	&96.22		&14.40 	&4.50 	&96.32        \\
DCTNet~\cite{zhao2022discrete}          &                             &31.56 	&12.16 	&91.84 	    &5.48 	&2.27 	&97.11 	    &10.69 	&4.30	&94.82		&15.91 	&6.24 	&94.59        \\
SUFT~\cite{shi2022symmetric}            &                             &26.17 	&6.80 	&96.25 	    &4.89 	&1.43 	&98.35 	    &8.74 	&2.47 	&97.18 		&13.27 	&3.57 	&97.26        \\
DADA~\cite{metzger2023guided}           &                             &32.86 	&12.51 	&91.34 	    &5.99 	&2.38 	&96.73 	    &12.31 	&4.96	&93.90		&17.05 	&6.62 	&93.99        \\
SGNet~\cite{wang2024sgnet}              &                             &25.32 	&6.58 	&96.32 	    &4.76 	&\underline{1.39} 	&98.40 	    &8.54 	&\underline{2.40} 	&\underline{97.22} 		&12.87 	&3.46 	&97.31        \\
DORNet~\cite{wang2025dornet}            &                             &28.39 	&8.39 	&95.05 	    &5.33 	&1.67 	&97.99 	    &9.93 	&3.21	&96.34		&14.55 	&4.42 	&96.46        \\
C2PD~\cite{kang2025c2pd}                &                             &\underline{23.76} 	&\underline{6.04} 	&\underline{96.50} 	    &\underline{4.71} 	&1.41 	&\underline{98.41} 	    &\underline{8.44} 	&2.42 	&97.17 		&\underline{12.30} 	&\underline{3.29} 	&\underline{97.36}        \\
SPFNet~\cite{wang2026scene}             &                             &25.83 	&6.50 	&96.44 	&4.91 	&1.42 	&98.38 	&8.78 	&2.44 	&97.20 	&13.17 	&3.45 	&97.34      \\
\rowcolor{gray!30}
\textbf{MOMNet-T}                       &                             &26.44 	&7.36 	&95.74 	    &\underline{4.71} 	&1.46 	&98.31 	    &8.93 	&2.68 	&96.90 		&13.36 	&3.83 	&96.98        \\
\rowcolor{gray!30}
\textbf{MOMNet}                         &                             &\textbf{22.99} 	&\textbf{5.24} 	&\textbf{97.24} 	    &\textbf{4.47} 	&\textbf{1.33} 	&\textbf{98.49} 	    &\textbf{7.89} 	&\textbf{2.17} 	&\textbf{97.49} 		&\textbf{11.78} 	&\textbf{2.91} 	&\textbf{97.74}        \\
\bottomrule
\end{tabular}}
\end{table*}
%%%%%%%%%%%%%%%%%%%%%%%%%%%%%%%%%%%%%%%%%%%%%%%%%%%%%%

\section{Experiments}
\label{sec:exp}

\subsection{Experimental Setups}
\begin{figure}[t]
\centering
\includegraphics[width=0.92\columnwidth]{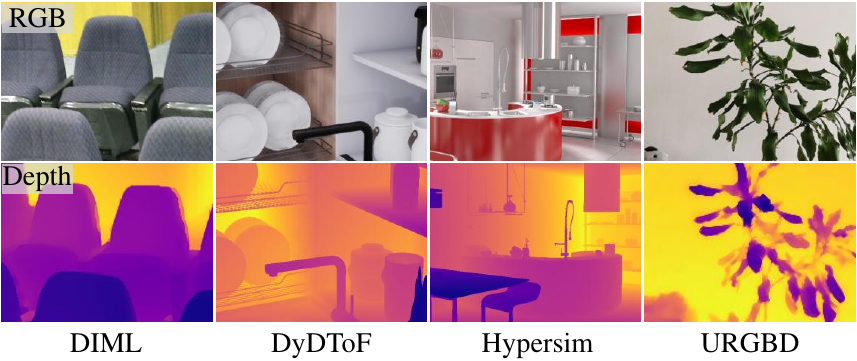}\\
\caption{Example visualization of unaligned RGB-D pairs.}\label{fig:examples}
\end{figure}

% \subsubsection{Datasets}
To thoroughly evaluate the robustness and generalization of MOMNet, we construct three synthetic misaligned datasets and one real-world misaligned test set, as illustrated in Fig.~\ref{fig:examples}.

%%%%%%%%%%%%%%%%%%%%%%% Fig. DatasetExample%%%%%%%%%%%%%%%%%%%%%%%%%%%
%%%%%%%%%%%%%%%%%%%%%%%%%%%%%%%%%%%%%%%%%%%%%%%%%%%%%%%%%%

\noindent \textbf{Synthetic Dataset.} We utilize three public datasets to synthesize non-aligned RGB-D data, including Hypersim \cite{roberts2021hypersim}, DIML \cite{cho2021diml, kim2017deep, kim2018deep, cho2021deep}, and DyDToF \cite{sun2023consistent}. Specifically, we introduce a cross-frame selection strategy that constructs misaligned RGB-D pairs by combining RGB and depth from different frames (\textit{e.g.}, pairing the RGB from frame $1$ with the depth from frame $4$). Owing to the random camera motion during the acquisition of depth and RGB videos, such cross‑frame pairing places RGB and depth under different motion trajectories, thereby producing misaligned data with diverse disparities, focal variations, as well as horizontal and vertical pixel shifts. We first apply this strategy to construct a misaligned Hypersim, comprising a training set of $1,500$ RGB-D pairs (a size similar to prior aligned DSR methods) with varying levels of misalignment and a test set of $100$ pairs. Then, the pre-trained weights from Hypersim are directly applied to test DIML ($100$ RGB-D pairs) and DyDToF ($100$ RGB-D pairs) datasets without any fine-tuning, thereby evaluating the generalization capability. The test sets of DIML, DyDToF, and Hypersim feature varying levels of misalignment, with approximately $10\%$, $20\%$, and $30\%$ of pixels misaligned, respectively.  Following previous methods~\cite{kim2021deformable, zhao2022discrete}, all LR depths are generated from GT depth using bicubic downsampling. Additionally, to ensure a fair comparison, we retrain all competing methods using the same datasets and settings as MOMNet.

%%%%%%%%%%%%%%%%%%%%%%% Fig. 7 -- DIML %%%%%%%%%%%%%%%%%%%%%%%%%%%
\begin{figure*}[t]
\centering
\includegraphics[width=0.88\linewidth]{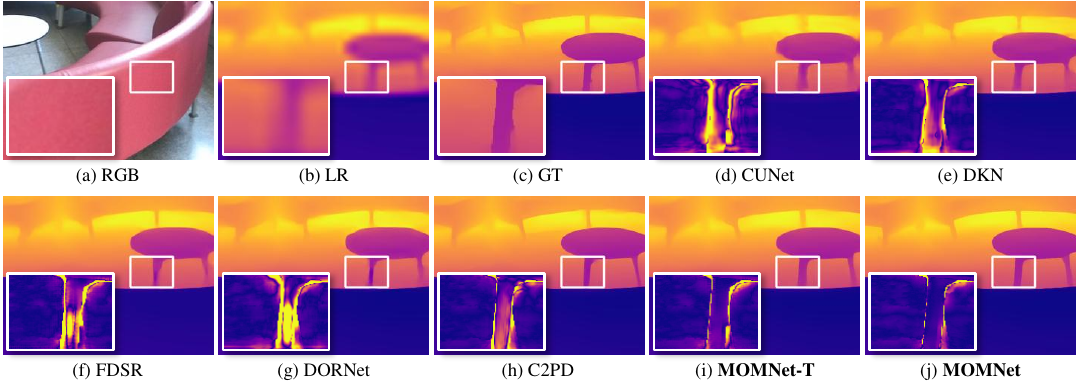}\\
\caption{Visual results and error maps (zoomed-in views of the white boxes in (d)-(j)) on $\times16$ DIML. Brighter colors in error maps denote larger errors.}\label{fig:diml_visual}
\end{figure*}
%%%%%%%%%%%%%%%%%%%%%%%%%%%%%%%%%%%%%%%%%%%%%%%%%%%%%%%%%%

%%%%%%%%%%%%%%%%%%%%%%% Fig. 8 -- URGBD %%%%%%%%%%%%%%%%%%%%%%%%%%%
\begin{figure*}[t]
\centering
\includegraphics[width=0.88\linewidth]{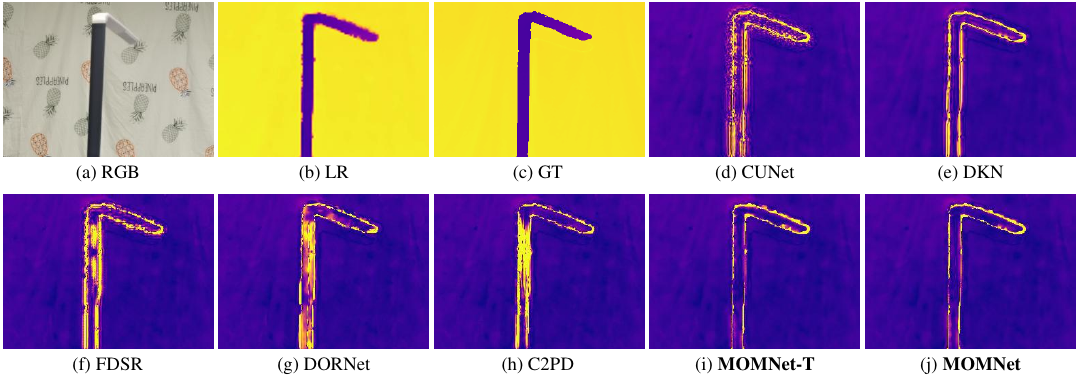}\\
\caption{Error comparison on the real-world unaligned URGBD (temporal asynchrony).}\label{fig:urgbd}
\end{figure*}
%%%%%%%%%%%%%%%%%%%%%%%%%%%%%%%%%%%%%%%%%%%%%%%%%%%%%%%%%%

%%%%%%%%%%%%%%%%%%%%%%%%%%% Tab. 2 Real URGBD %%%%%%%%%%%%%%%%%%%%%%%%%%%
\begin{table*}[t]
\caption{Quantitative comparisons on real-world URGBD dataset.}\label{tab:real_NON}
	\centering
	\resizebox{1\linewidth}{!}{
\begin{tabular}{c|cccccccc>{\columncolor{gray!30}}c>{\columncolor{gray!30}}c}
\toprule 
Metrics  &DJFR~\cite{li2019joint} &CUNet~\cite{deng2020deep} & DKN~\cite{kim2021deformable}   &FDSR~\cite{he2021towards}   &DCTNet~\cite{zhao2022discrete}  &DORNet~\cite{wang2025dornet} &C2PD~\cite{kang2025c2pd} &SPFNet~\cite{wang2026scene}   &\textbf{MOMNet-T}&\textbf{MOMNet}\\
\midrule
RMSE$\downarrow $                  	     &8.60     &8.63         &8.77     &9.07       &8.86     &9.11     &8.93                &8.57   &\textbf{8.35}     &\underline{8.39}  \\
MAE$\downarrow $               	         &4.12     &4.45         &4.10     &4.22       &4.53     &4.24     &4.17                &4.11   &\underline{4.08}     &\textbf{4.05}  \\
$\delta _{1.05} \uparrow $     	         &92.49    &91.82        &92.44    &92.19      &92.37    &92.64    &\underline{92.66}   &92.60   &92.56              &\textbf{92.68}  \\
\bottomrule
\end{tabular}}
\end{table*}
%%%%%%%%%%%%%%%%%%%%%%%%%%%%%%%%%%%%%%%%%%%%%%%

\noindent \textbf{Real-World Dataset.} To evaluate generalization in real‑world scenarios, we construct URGBD, a real‑world unaligned dataset captured using unregistered RGB and depth cameras. It contains $50$ RGB‑D pairs exhibiting varying degrees of misalignment arising from differences in focal length, parallax, and temporal asynchrony. Moreover, the RGB and LR depth in URGBD are acquired by a real color camera and a real Time‑of‑Flight (ToF) sensor, respectively.

\noindent \textbf{Implementation Details.} We select root mean square error (RMSE) in centimeters, mean absolute error (MAE) in centimeters, and $\delta _{1.05}$ as evaluation metrics. The Adam optimizer with an initial learning rate of $1\times 10^{-4} $ is used to train our approach. Additionally, we implement MOMNet on a single NVIDIA 4090 GPU. The hyperparameter $\alpha$ is set to $0.001$.

\subsection{Comparison with the State-of-the-Art}
\noindent \textbf{Quantitative Comparison.} As shown in Tab.~\ref{tab:qua}, MOMNet demonstrates robust performance across various levels of misalignment, scale factors, and metrics. Overall, the performance advantage of our method becomes increasingly evident as the misalignment level rises. For example, compared to the suboptimal method C2PD~\cite{kang2025c2pd}, our MOMNet reduces RMSE (averaged over three scales) by $0.19cm$, $0.46cm$, and $1.28cm$ under $\sim10\%$, $\sim20\%$, and $\sim30\%$ misalignment, respectively. Moreover, our lightweight MOMNet-T also delivers competitive performance, with an average RMSE ($\times16$) advantage of $2.55cm$ over DCTNet~\cite{zhao2022discrete} and $1.19cm$ over DORNet~\cite{wang2025dornet}.

%%%%%%%%%%%%%%%%%%%%%%%% Tab. 3 --fine-tuning results %%%%%%%%%%%%%%%%%%%%%%
\begin{table*}[t]
\caption{Quantitative comparisons of fine-tuning results with existing methods.}\label{tab:fine}
% \vspace{-4pt}
\centering
% \footnotesize
\resizebox{0.95\linewidth}{!}{
\begin{tabular}{l|c|cccccccccccc}
\toprule 
\multirow{2}{*}{Methods} &\multirow{2}{*}{Scale}   &\multicolumn{3}{c}{Hypersim ($\sim30\%$)}   &\multicolumn{3}{c}{DIML ($\sim10\%$)}  &\multicolumn{3}{c}{DyDToF ($\sim20\%$)}   &\multicolumn{3}{c}{Average}\\ %\cline {3-14}
%\cmidrule{2-13}
\cmidrule(lr){3-5}\cmidrule(lr){6-8}\cmidrule(lr){9-11}\cmidrule(lr){12-14} 
 & &RMSE$\downarrow $ &MAE$\downarrow $ &$\delta _{1.05} \uparrow $     &RMSE$\downarrow $ &MAE$\downarrow $ &$\delta _{1.05} \uparrow $  
 &RMSE$\downarrow $ &MAE$\downarrow $ &$\delta _{1.05} \uparrow $     &RMSE$\downarrow $ &MAE$\downarrow $ &$\delta _{1.05} \uparrow $ \\ \midrule
DCTNet~\cite{zhao2022discrete}          &\multirow{7}{*}{$\times 8$}  &19.55 	&5.41 	&96.81 	        &2.89 	&0.86 	&99.27 	    &5.97 	&1.75 	&98.03 		&9.47 	&2.67 	&98.04       \\
SGNet~\cite{wang2024sgnet}              &                             &\underline{14.38} 	&\underline{2.28} 	&\underline{98.97} 	        &2.82 	&\textbf{0.62} 	&\underline{99.37} 	    &\underline{4.80} 	&0.93 	&98.93 		&\underline{7.33} 	&\underline{1.28} 	&\underline{99.09}       \\
DORNet~\cite{wang2025dornet}            &                             &17.39 	&3.10 	&98.53 	        &3.10 	&0.67 	&99.32 	    &5.58 	&1.16 	&98.65 		&8.69 	&1.64 	&98.83       \\
C2PD~\cite{kang2025c2pd}                &                             &14.73 	&2.42 	&98.87 	        &\underline{2.81} 	&\textbf{0.62} 	&\textbf{99.38} 	    &4.82 	&0.98 	&98.92 		&7.45 	&1.34 	&99.06       \\
SPFNet~\cite{wang2026scene}             &                             &15.69 	&2.45 	&98.91 	&2.97 	&\underline{0.63}	&99.36 	&4.84 	&\underline{0.90} 	&\underline{99.01} 	&7.83 	&1.33 	&\underline{99.09}      \\
\rowcolor{gray!30}
\textbf{MOMNet-T}                       &                             &17.52 	&3.35 	&98.28 	&\textbf{2.76} 	&\underline{0.63} 	&\underline{99.37} 	    &5.47 	&1.21 	&98.55 		&8.58 	&1.73 	&98.73      \\
\rowcolor{gray!30}
\textbf{MOMNet}                         &                             &\textbf{13.91} 	&\textbf{2.09} 	&\textbf{99.06} 	&\underline{2.81} 	&\textbf{0.62} 	&\textbf{99.38} 	    &\textbf{4.41} 	&\textbf{0.83} 	&\textbf{99.09} 		&\textbf{7.04} 	&\textbf{1.18} 	&\textbf{99.18}      \\
\midrule       
DCTNet~\cite{zhao2022discrete}          &\multirow{7}{*}{$\times 16$} &29.72 	&10.92 	&92.86 	        &5.10 	&2.01 	&97.63 	    &10.18 	&3.93 	&95.44 		&15.00 	&5.62 	&95.31       \\
SGNet~\cite{wang2024sgnet}              &                             &24.53 	&6.15 	&96.60 	        &4.67 	&\underline{1.38} 	&98.45 	    &8.37 	&2.37 	&97.24 		&12.52 	&3.30 	&97.43       \\
DORNet~\cite{wang2025dornet}            &                             &27.65 	&7.94 	&95.34 	        &5.16 	&1.62 	&98.07 	    &9.73 	&3.06 	&96.53 		&14.18 	&4.21 	&96.65       \\
C2PD~\cite{kang2025c2pd}                &                             &\underline{23.49} 	&\underline{5.75} 	&\underline{96.92} 	        &\underline{4.62} 	&1.39 	&\underline{98.46} 	    &\underline{8.28} 	&\underline{2.33} 	&\underline{97.32} 		&\underline{12.13} 	&\underline{3.16} 	&\underline{97.57}       \\
SPFNet~\cite{wang2026scene}             &                             &25.53 	&6.35 	&96.51 	&4.86 	&\underline{1.38} 	&98.43 	&8.49 	&\underline{2.33} 	&97.31 	&12.96 	&3.35 	&97.42      \\
\rowcolor{gray!30}
\textbf{MOMNet-T}                       &                             &26.44 	&7.36 	&95.74 	    &4.71 	&1.46 	&98.31 	    &8.93 	&2.68 	&96.90 		&13.36 	&3.83 	&96.98        \\
\rowcolor{gray!30}
\textbf{MOMNet}                         &                             &\textbf{22.99} 	&\textbf{5.24} 	&\textbf{97.24} 	    &\textbf{4.47} 	&\textbf{1.33} 	&\textbf{98.49} 	    &\textbf{7.89} 	&\textbf{2.17} 	&\textbf{97.48} 		&\textbf{11.78} 	&\textbf{2.91} 	&\textbf{97.74}        \\
\bottomrule
\end{tabular}}
\end{table*}
%%%%%%%%%%%%%%%%%%%%%%%%%%%%%%%%%%%%%%%%%%%%%%%%%%%%%%

To assess the generalization of our approach in real-world scenarios, we apply the model pre-trained on the Hypersim dataset directly to the real-world URGBD dataset without any fine-tuning. Tab.~\ref{tab:real_NON} demonstrates that our method exhibits strong robustness in real-world misaligned scenes, achieving a $ 0.22cm$ improvement in RMSE over the suboptimal method.

%%%%%%%%%%%%%%%%%%%%%%% Fig.9 Params_Time  %%%%%%%%%%%%%%%%%%%%%%%%%%%
\begin{figure}[t]
\centering
\includegraphics[width=0.92\columnwidth]{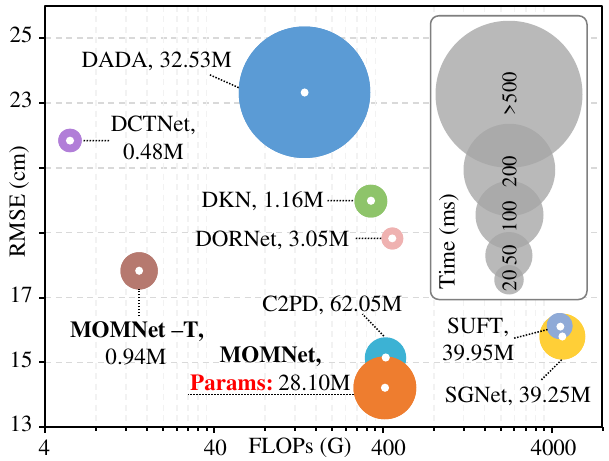}\\
\caption{Complexity comparison on $\times8$ Hypersim tested by a 4090
GPU. A larger circle area indicates longer inference time.}\label{fig:params}
\end{figure}
%%%%%%%%%%%%%%%%%%%%%%%%%%%%%%%%%%%%%%%%%%%%%%%%%%%%%%%%%%

%%%%%%%%%%%%%%%%%%%%%%% Fig.10 NoiseRobustness  %%%%%%%%%%%%%%%%%%%%%%%%%%%
\begin{figure}[t]
\centering
\includegraphics[width=0.98\columnwidth]{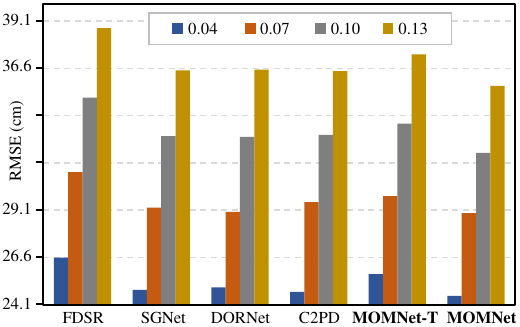}\\
\caption{Robustness to different gaussian noise (standard deviation ranging from $0.04$ to $0.13$) on the $\times8$ Hypersim dataset.}\label{fig:noise}
\end{figure}
%%%%%%%%%%%%%%%%%%%%%%%%%%%%%%%%%%%%%%%%%%%%%%%%%%%%%%%%%%

\noindent \textbf{Visual Comparison.} Fig.~\ref{fig:diml_visual} presents a visual comparison with state-of-the-art methods on the DIML dataset, demonstrating that our approach effectively reconstructs high-quality depth from misaligned RGB and LR depth inputs. It is evident that the depth of the table legs predicted by our MOMNet is closer to GT, with fewer errors. Furthermore,  Fig.~\ref{fig:urgbd} shows a comparison of error maps on the real-world unaligned URGBD. It can be clearly observed that the depth reconstructed by our MOMNet and MOMNet-T exhibit substantially fewer errors compared to those produced by other methods.

\noindent \textbf{Comparison of Fine-Tuning Results with Prior Methods.} To ensure a fair and comprehensive evaluation, we additionally fine-tuned previous methods on the misaligned dataset using their publicly available checkpoints. As shown in Tab.~\ref{tab:fine}, our method still achieves competitive performance and outperforms these fine-tuned results. For example, compared to the suboptimal method, our MOMNet reduces RMSE by $0.50cm$ and MAE by $0.51cm$, while increasing $\delta_{1.05}$ by $0.32$ on the $\times16$ Hypersim dataset. These results fully demonstrate the effectiveness and robustness of our method.

\noindent \textbf{Model Complexity Analysis.} As shown in Fig.~\ref{fig:params}, our method achieves a competitive balance between performance and model complexity. Compared to the suboptimal approach C2PD~\cite{kang2025c2pd}, our MOMNet exhibits comparable computational cost while achieving a significant performance improvement, reducing the parameter count by $33.95M$ and further lowering the RMSE by $0.92cm$. For lightweight DSR, MOMNet-T also delivers satisfactory performance. It surpasses the latest DORNet~\cite{wang2025dornet} by achieving an additional performance improvement of $1.00cm$, while reducing the number of parameters by $2.11M$ and FLOPs by $440.22G$.

\begin{table*}[t]
\caption{Quantitative comparisons on the real-world aligned RGB-D-D and TOFDSR.}\label{tab:real_aligned}
	\centering
	% \Large
	\resizebox{0.97\linewidth}{!}{
\begin{tabular}{c|c|ccccccc>{\columncolor{gray!30}}c>{\columncolor{gray!30}}c}
\toprule 
Datasets &Metrics   &DJFR~\cite{li2019joint} &CUNet~\cite{deng2020deep} & DKN~\cite{kim2021deformable}   &FDSR~\cite{he2021towards}   &DCTNet~\cite{zhao2022discrete}   &SGNet~\cite{wang2024sgnet} &SPFNet~\cite{wang2026scene}  &\textbf{MOMNet-T}&\textbf{MOMNet}\\
\midrule
\multirow{3}{*}{RGB-D-D} 
&RMSE$\downarrow $                  	     &5.52     &5.84         &5.08     &5.49       &5.43          &5.42    &\underline{4.21}   &4.32     &\textbf{4.18}  \\
&MAE$\downarrow $               	         &3.51     &3.06         &2.58     &3.10       &3.29          &3.21    &1.87   &\underline{1.53}     &\textbf{1.40}  \\
&$\delta _{1.05} \uparrow $     	        &93.58    &94.75        &96.28    &94.77      &93.15          &94.77   &\underline{97.22}   &97.17    &\textbf{97.41}  \\
\midrule
\multirow{3}{*}{TOFDSR} 
&RMSE$\downarrow $                  	     &5.72       &6.04         &5.50     &5.03       &5.16          &\textbf{4.33}       &4.58   &4.73     &\underline{4.40}  \\
&MAE$\downarrow $               	         &2.10       &2.21         &2.07     &1.67       &2.10          &1.33                &1.33   &\underline{1.23}     &\textbf{1.03}  \\
&$\delta _{1.05} \uparrow $     	         &97.03      &96.46        &97.54    &97.61      &96.37         &98.35               &\underline{98.39}   &97.90     &\textbf{98.46}  \\
\bottomrule
\end{tabular}}
\end{table*}
%%%%%%%%%%%%%%%%%%%%%%%%%%%%%%%%%%%%%%%%%%%%%%%

\noindent \textbf{Robustness to Noise.} To further validate the robustness of MOMNet, Fig.~\ref{fig:noise} presents a quantitative comparison across different noise levels. Similar to previous approaches~\cite{kim2021deformable, wang2025dornet}, we add Gaussian noise (mean $0$, standard deviation ranging from $0.04$ to $0.13$) to the LR depth as a new input. These results indicate that our method achieves excellent noise robustness, consistently attaining the lowest RMSE under all noise levels. For example, MOMNet achieves an RMSE reduction of $0.85cm$ and $0.77cm$ over the suboptimal method at standard deviations of $0.10$ and $0.13$, respectively. 

\noindent \textbf{Comparison in Real-World Aligned Scenarios.} Following the experimental protocols of previous methods~\cite{he2021towards,wang2024sgnet}, we train and evaluate our approach on the real-world aligned RGB-D-D and TOFDSR. As reported in Tab.~\ref{tab:real_aligned}, the quantitative results confirm that our approach maintains strong robustness in aligned scenarios. For example, compared to the suboptimal method, our MOMNet reduces the MAE by $25.13\%$ on the TOFDSR dataset and by $22.56\%$ on the RGB-D-D dataset. These results further demonstrate  the effectiveness of our method for real-world scenes.

%%%%%%%%%%%%%%%%%%%%%%%%%%%%Table.2 Ablation1%%%%%%%%%%%%%%%%%%%%%%%%%%
\begin{table}[t]
\caption{Ablation studies of zero-order matching (ZOM), first-order matching (FOM), and second-order matching (SOM) on $\times8$ DSR.} \label{tab:ab_1}
	\centering
 \large
	    \resizebox{1\linewidth}{!}{ 
		\begin{tabular}{c|ccccccc}
			\toprule 
				Methods   	 &ZOM          &FOM 	     &SOM      &Hypersim     &DyDToF  &Params.   &FLOPs 	\\ %\midrule
			\midrule
                (a)          &             & 	         &         &15.31 	           &4.66  &12.11M    &222.64G \\
				(b)          &$\surd$      & 	         &         &14.51 	           &4.62  &15.76M    &260.84G   \\
                (c)          &             &$\surd$ 	 &         &14.83 	           &4.52  &16.18M    &267.42G   \\
				(d)          &             & 	         &$\surd$  &14.67 	           &4.57  &16.18M    &267.42G   \\
                (e)          &$\surd$      &$\surd$ 	 &         &14.28 	           &4.48  &23.16M    &351.48G   \\
                (f)          &$\surd$      & 	         &$\surd$  &14.39 	           &4.51  &23.16M    &351.48G   \\
				(g)          &             &$\surd$ 	 &$\surd$  &14.51 	           &4.51  &21.11M    &326.31G   \\
                \rowcolor{gray!30}
                (h)          &$\surd$      &$\surd$ 	 &$\surd$  &\textbf{14.14} 	           &\textbf{4.41}   &28.10M    &410.38G  \\
			\bottomrule  
			%\bottomrule
		\end{tabular}}
\end{table}
%%%%%%%%%%%%%%%%%%%%%%%%%%%%%%%%%%%%%%%%%%%%%%%%%%%%%%

\subsection{Ablation Studies}

\noindent \textbf{Multi-Order Matching.} Tab.~\ref{tab:ab_1} presents the ablation study of zero-order, first-order, and second-order matching. For the baseline (a), we remove all matching modules and use the original RGB input to replace all matched RGB features. Compared with (a), (b)–(d) demonstrate that each individual matching algorithm consistently improves performance while incurring only a modest computational overhead. Furthermore, (e)–(g) validate the complementarity of multi-order matching, showing that it is more effective to enhance reconstruction quality than any single matching. When all three are combined, our method (h) achieves the best performance, reducing RMSE by $1.17cm$ on Hypersim compared with the baseline. Overall, these results fully confirm the effectiveness and synergistic benefits of our multi-order matching strategy.

%%%%%%%%%%%%%%%%%%%%%%% Fig.11 Ab_MOM_MOA  %%%%%%%%%%%%%%%%%%%%%%%%%%%
\begin{figure}[t]
\centering
\includegraphics[width=0.97\columnwidth]{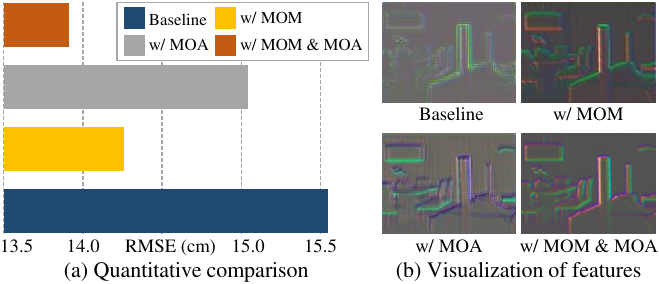}\\
\caption{Ablation study of multi-order matching (MOM) and multi-order aggregation (MOA) on $\times8$ Hypersim dataset.}\label{fig:ab_MOM_MOA}
\end{figure}
%%%%%%%%%%%%%%%%%%%%%%%%%%%%%%%%%%%%%%%%%%%%%%%%%%%%%%%%%%

%%%%%%%%%%%%%%%%%%%%%%% Fig.12 ab_MOMA_Numbers_SD  %%%%%%%%%%%%%%%%%%%%%%%%%%%
\begin{figure}[t]
\centering
\includegraphics[width=0.99\columnwidth]{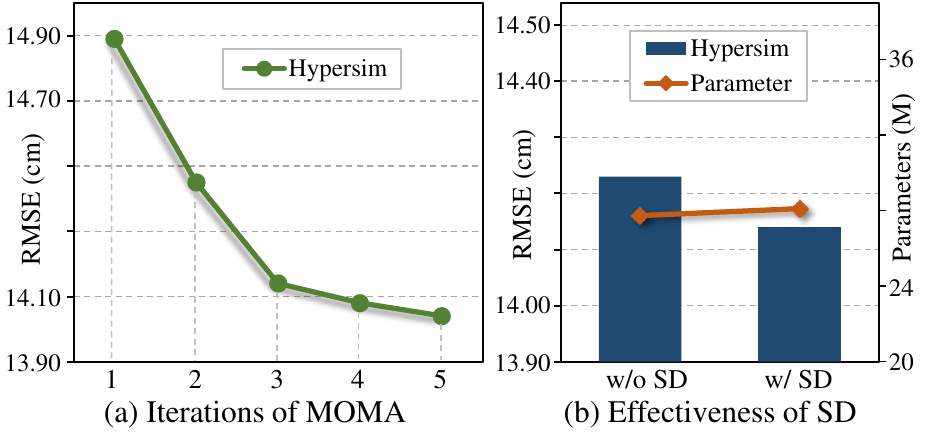}\\
\caption{Ablation study of MOMNet with (a) MOMA numbers and (b) effectiveness of structure detector (SD) on $\times8$ Hypersim dataset.}\label{fig:ab_MOMA_Numbers_SD}
\end{figure}
%%%%%%%%%%%%%%%%%%%%%%%%%%%%%%%%%%%%%%%%%%%%%%%%%%%%%%%%%%

\noindent \textbf{MOMA.} Fig.~\ref{fig:ab_MOM_MOA}(a) depicts the ablation study of multi-order matching (MOM) and multi-order aggregation (MOA). For the baseline, we remove all MOM and MOA from MOMNet and use addition to fuse the misaligned RGB-D. The results clearly show that both MOM and MOA contribute to performance gains over the baseline. When combined, our method achieves the best results, surpassing the baseline by $1.64cm$ RMSE. Additionally, Fig.~\ref{fig:ab_MOM_MOA}(b) visualizes the depth features using principal component analysis (PCA), showing that both MOM and MOA effectively enhance the depth structure.

Fig.~\ref{fig:ab_MOMA_Numbers_SD}(a) presents the results with different numbers of multi-order matching and aggregation (MOMA). It can be seen that the error progressively decreases as the number of iterations increases. When the number exceeds $3$, the declining trend of RMSE gradually plateaus. To balance model complexity and performance, we adopt $3$ as the default setting.

\noindent \textbf{Structure Detector.} Fig.~\ref{fig:ab_MOMA_Numbers_SD}(b) reports the ablation study of the structure detector (SD). These results demonstrate that SD effectively contributes to performance improvement with only a minor increase in parameters. For instance, SD successfully reduces the RMSE by $0.08cm$ on the Hypersim dataset while introducing only $0.38M$ additional parameters.

\noindent \textbf{Different Loss Functions.} Tab.~\ref{tab:ab_loss} lists the ablation study of different loss functions. The baseline model removes both the first-order gradient term $ \mathcal{L}_{grad}$ and the second-order Hessian term $ \mathcal{L}_{hes}$ from the multi-order regularization, retaining only the zero-order reconstruction term $\mathcal{L}_{rec} $. Compared to the baseline, incorporating either $ \mathcal{L}_{grad}$ or $ \mathcal{L}_{hes}$ individually leads to performance gains. When all three terms are combined, our MOMNet achieves the best performance, surpassing the baseline by $0.13cm$ in RMSE on the Hypersim dataset.

%%%%%%%%%%%%%%%%%%%%%%%%%%%%Table.3 Ablation2%%%%%%%%%%%%%%%%%%%%%%%%%%
\begin{table}[t]
\caption{Ablation studies of different loss functions on $\times8$ DSR.} 
\label{tab:ab_loss}
	\centering
 \small
	    \resizebox{0.7\linewidth}{!}{ 
		\begin{tabular}{l|cc}
			\toprule 
				Methods   	                                                 &Hypersim          &DyDToF \\ 
			\midrule
				Baseline                                                     &14.25      &4.47 	          \\
                Baseline + $\mathcal{L}_{grad}$          &14.16             &4.44    \\
				Baseline + $\mathcal{L}_{hes}$          &14.22             &4.46 	          \\
                \rowcolor{gray!30}
                Baseline + $\mathcal{L}_{grad}$ + $\mathcal{L}_{hes}$          &\textbf{14.14} 	           &\textbf{4.41} 	 \\
			\bottomrule  
			%\bottomrule
		\end{tabular}}
\end{table}
%%%%%%%%%%%%%%%%%%%%%%%%%%%%%%%%%%%%%%%%%%%%%%%%%%%%%%

%%%%%%%%%%%%%%%%%%%%%%% Fig.13 TOPK  %%%%%%%%%%%%%%%%%%%%%%%%%%%
\begin{figure}[t]
\centering
\includegraphics[width=1\columnwidth]{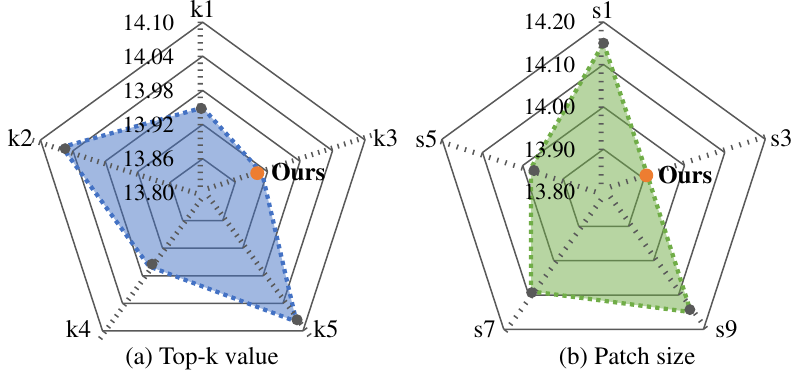}\\
\caption{Ablation study of MOMNet with (a) different top-k values and (b) different patch sizes on $\times8$ Hypersim. `k1' selects the top-$1$ matched RGB patch, and `s1' uses $1\times1$ patch for matching retrieval.}\label{fig:ab_topk}
\end{figure}
%%%%%%%%%%%%%%%%%%%%%%%%%%%%%%%%%%%%%%%%%%%%%%%%%%%%%%%%%%

\noindent \textbf{Matching Retrieval.} 
Fig.~\ref{fig:ab_topk}(a) illustrates the ablation study on different top‑$k$ values in the matching retrieval. Specifically, we conduct five sets of experiments with $k$ values ranging from $1$ to $5$. The results reveal that the error does not consistently decrease as $k$ increases. Our method achieves the best performance when $k=3$. For instance, `k3' outperforms `k2' by $0.15cm$ and surpasses `k5' by $0.17cm$. Therefore, we adopt $3$ as the default value for $k$. Besides, Fig.~\ref{fig:ab_topk}(b) presents the ablation study of five different patch sizes in the matching retrieval. We find that performance does not vary monotonically with the patch size. For example, `s5' contributes better performance than `s7', while `s9' outperforms `s5'.  Among all configurations, the `s3' achieves the best performance.  Consequently, we adopt `s3' as the default setting.

\section{Conclusion}
In this paper, we propose MOMNet, a novel paradigm for alignment-free DSR. To alleviate the reliance on strictly spatially aligned RGB-D data, our method first collaboratively identifies RGB content relevant to depth by performing multi-order matching across zero-order, first-order, and second-order feature spaces. Based on multi-order priors, we further design a multi-order aggregation strategy that dynamically transfers the matched RGB information to depth representation using the structure detector. Additionally, a multi-order regularization is introduced to facilitate the optimization of our MOMNet, thereby enhancing geometric consistency in the predicted depth. Extensive experiments on both unaligned and aligned data demonstrate the robustness of our MOMNet.

\bibliographystyle{IEEEtran}
\bibliography{reference}

\end{document}